\documentclass[conference, a4paper]{IEEEtran}

\usepackage{siunitx}
\usepackage[T1]{fontenc}
\usepackage{float}
\usepackage[utf8]{inputenc}
\usepackage[square, numbers, comma, sort & compress]{natbib}
\usepackage{amsthm}      
\usepackage{amsmath}
\usepackage{makecell, cellspace, caption}
\usepackage{flushend}
\usepackage{amssymb}
\usepackage{subcaption}
\usepackage{acronym} 
\usepackage{mathrsfs}
\usepackage{booktabs}
\usepackage{url}
\usepackage{algorithm}
\usepackage[noend]{algpseudocode}
\usepackage{multirow}
\usepackage[shortcuts,acronym]{glossaries}
\usepackage{mathtools}
\usepackage{afterpage} 
\usepackage[normalem]{ulem}
\usepackage{balance}
\usepackage{pgfplots}
\usepackage{array}
\usepackage{threeparttable}
\usepackage{comment}
\usepackage{pifont}
\usepackage{multicol}
\usepackage{wrapfig}
\usepackage{colortbl}

\newcolumntype{M}[1]{>{\centering\arraybackslash}m{#1}}
\newcolumntype{g}{>{\columncolor{Gray}}c}

\definecolor{gray9}{gray}{.9}
\definecolor{gray95}{gray}{.95}
\definecolor{gray8}{gray}{.8}
\definecolor{gray85}{gray}{.85}

\pgfplotsset{compat=newest}

\newacronym{cavs}{CAVs}{Connected and Autonomous Vehicles}
\newacronym{v2x}{V2X}{Vehicle-to-Everything}
\newacronym{cpm}{CPM}{Cooperative Perception Message}
\newacronym{etsi}{ETSI}{The European Telecommunications Standards Institute}
\newacronym{v8}{YOLOv8}{YOLOv8}
\newacronym{bev}{BEV}{Bird's-eye view}
\newacronym{nds}{NDS}{NuScenes Detection Score}
\newacronym{cps}{CPS}{Cooperative Perception System}

\begin{document}

\title{Leveraging V2X for Collaborative HD Maps Construction Using Scene Graph Generation}

\author{
    \IEEEauthorblockN{Gamal Elghazaly and Raphael Frank}
    \IEEEauthorblockA{
        Interdisciplinary Center for Security, Reliability and Trust (SnT), University of Luxembourg\\
        29 Avenue J.F Kennedy, L-1855 Luxembourg\\
        firstname.lastname@uni.lu
    }
}
\maketitle

\begin{abstract}

High-Definition (HD) maps play a crucial role in autonomous vehicle navigation, complementing onboard perception sensors for improved accuracy and safety. Traditional HD map generation relies on dedicated mapping vehicles, which are costly and fail to capture real-time infrastructure changes. This paper presents \textit{HDMapLaneNet}, a novel framework leveraging V2X communication and Scene Graph Generation to collaboratively construct a localized geometric layer of HD maps. The approach extracts lane centerlines from front-facing camera images, represents them as graphs, and transmits the data for global aggregation to the cloud via V2X. Preliminary results on the nuScenes dataset demonstrate superior association prediction performance compared to a state-of-the-art method.
\end{abstract}
\vspace{5px}

\section{Introduction}
\label{sec:introduction}

High-Definition (HD) maps provide critical information for autonomous driving, complementing onboard perception sensors to ensure safe and reliable navigation~\cite{testouri2024robocarrapidlydeployableopensource}. The creation of such detailed maps typically requires dedicated mapping vehicles that periodically scan road environments to generate the various layers of an HD map, which are then made available to autonomous vehicles and mobility services~\cite{elghazaly2023high}. However, this approach is costly and does not always capture real-time changes in road infrastructure (e.g., due to construction).

To address these limitations, alternative techniques leveraging collaborative sensing by autonomous robots and vehicles have been explored. Recent research has investigated methods to dynamically sense driving environments, providing more accurate and timely information about moving agents~\cite{hawlader2024cooperative} and static road features to locally generate HD maps~\cite{dias2023hd} or construct them globally via crowdsourcing~\cite{9357917}.

In this paper, we present preliminary results of \textit{HDMapLaneNet}, a novel framework for generating localized portions of the geometric layer of an HD map using Scene Graph Generation. This technique can be applied by individual vehicles or robots, requiring only images from a front-facing camera to generate a graph representing lane centerlines. The generated graph is then converted into an HD map format and transmitted to the cloud via a V2X communication interface, where local maps are merged to form the final HD map’s geometric layer.

The method relies on a pipeline of interconnected neural networks and is validated on the nuScenes dataset~\cite{nuscenes}, which includes ground-truth HD maps. Our preliminary results demonstrate that the proposed approach outperforms a state-of-the-art method in association prediction.

The remainder of this paper is structured as follows: Section II details the proposed framework, Section III presents the implementation and results, and Section IV concludes with future research directions.

\section{Framework}

\begin{figure*}[t!]
    \centering  
    \includegraphics[scale=0.5, trim= 11cm 3.2cm 7cmcm 1.1cm,clip]{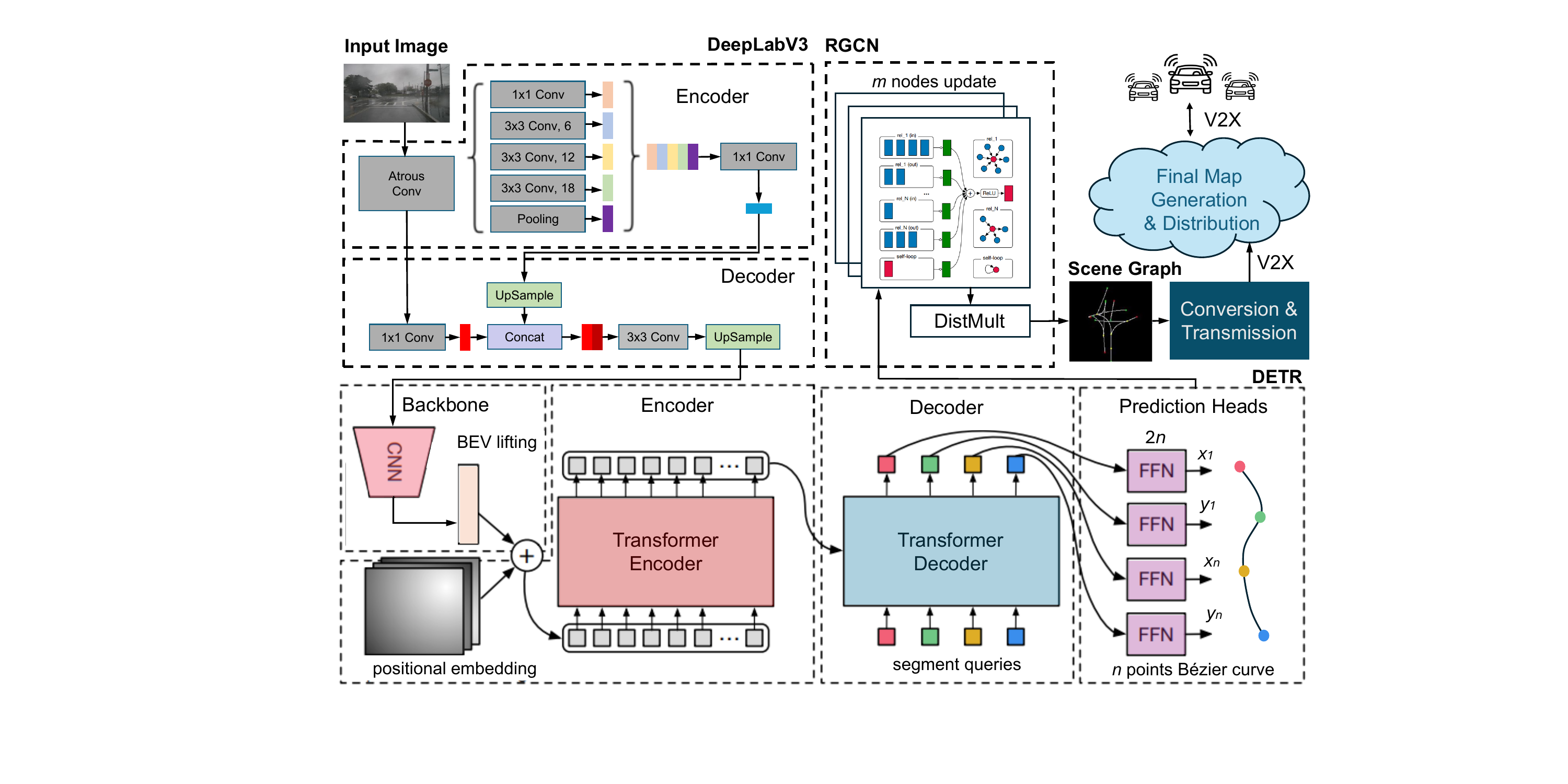}
    \caption{The high-level architecture of HDMapLaneNet. The pipeline begins by processing camera images using DeepLabV3 \cite{deeplabv3} as an image-view feature extractor. Simultaneously, lane centerline segments are detected using DETR \cite{detr}, which represents them as a Bézier curve. An RGCN \cite{RCGN} then constructs a scene graph by modeling the connectivity between segments. Finally, the serialized scene graph is transmitted via V2X communication for aggregation and distribution.}
    \label{fig:arch}
\end{figure*}

The proposed framework constructs a localized geometric layer of an HD Map. It represents lane centerlines as directed graphs and integrates deep learning models, including Convolutional Neural Networks (CNN) and transformers, for feature extraction and connectivity detection. The processed lane data are converted into a geo-referenced format and transmitted to the cloud, via a V2X communication interface. More details on the road representation and architecture will be given in the following subsections.

\subsection{Lane Representation}
In this work, we use directed graphs to represent lane centerlines. It is particularly helpful to provide an organized and structured representation of lane segments and define how they are connected. In this context, a single directed graph \( G=(V, E, R) \) can be used to represent the set of centerlines contained in an HD map. Graph vertices V correspond to road centerlines, and are represented using Bézier curves.
Given a set of $n$ lane control points \(S=\{P_1, P_2, ...., P_n\}\), a Bézier curve maps scalars \(t\in [0,1]\)  to 2D points in \(\mathbb{R}^2\) using the parametric representation defined by:

\begin{equation}
B(t)= \sum_{k=0}^{n} \binom{n}{k} (1-t)^{n-k} t^k P_k
\end{equation}

Detecting a centerline of a lane amounts to determining the control points of a Bézier curve. This representation is particularly useful to model any lane whatever its length using a fixed $n$ number of control points.
Graph edges \(E\subseteq \{(x,y)\mid (x,y) \in V^2 \wedge x \ne y \}\), on the other hand, correspond to lane centerlines connectivity and are represented using an incidence matrix $I$. Given a set of $m$ lane segments, the incidence matrix \( I_{m \times m} \) elements are defined by $1$ if two vertices are connected and $0$ otherwise.

\subsection{Architecture}

The proposed overall architecture for constructing HD map geometric lane segments and transmitting them to the cloud via a V2X communication interface is depicted in Fig. \ref{fig:arch}. The backbone of the architecture receives raw images from a front-facing camera and extracts map features using various processing blocks, which will be described in detail in the following subsections. The final step involves transforming the segments into an HD map-compatible format and sending them to the cloud for final map generation and redistribution.

\subsubsection{DeepLabv3}

DeepLabv3~\cite{deeplabv3, chen2017rethinkingatrousconvolutionsemantic} is an image segmentation model, whose encoder is based primarily on a CNN utilizing atrous (dilated) convolutions \cite{chen2017rethinkingatrousconvolutionsemantic}. This design allows it to incorporate commonly used classification architectures such as ResNet. To capture multi-scale contextual information, the encoder employs Atrous Spatial Pyramid Pooling (ASPP), which applies atrous convolutions with different dilation rates to extract features at multiple scales efficiently~\cite{deeplabv3, chen2017rethinkingatrousconvolutionsemantic}. The decoder module further refines segmentation accuracy by integrating low-level spatial details with high-level semantic features, leading to improved boundary delineation.

\subsubsection{Detection Transformer}
 
The Detection Transformer (DETR)~\cite{detr} is the first transformer-based object detection model, eliminating the need for complex heuristics and specialized layers by framing detection as a set prediction problem, enabling an end-to-end architecture with simplified processing and high accuracy.

The model consists of three main components: (1) a CNN backbone, (2) an encoder-decoder transformer, (3) and a feedforward network.

The Backbone extracts low-resolution feature maps from the input image. Its output is transformed into a one-dimensional feature map, which is then combined with positional encodings before being fed into the transformer encoder. We use the same positional encodings as in \cite{STSU} to incorporate Bird's Eye View (BEV) spatial information.

The Encoder-Decoder Transformer consists of multiple layers, each comprising a self-attention module and a feedforward network. The output of this block is a set of $K$ embeddings of fixed length, representing the number of objects the model assumes to be present in the image.

The Feedforward Network (FFN) consists of a three-layer perceptron with ReLU activation followed by a linear projection layer. It predicts the $N$ Bézier curve control points. At this stage, the graph vertices $V$ are encoded using Bézier curves, where each centerline is represented by a fixed-length vector containing 2D coordinates of the Bézier control points.

\subsubsection{Relational Graph Convolutional Network}

The next block is the Relational Graph Convolutional Network (RGCN)~\cite{RCGN}, which is used to predict associations between estimated lane centerline segments. Given a graph \( G = (V, E, R) \), where \( v_i \in V \) represents a graph node, \( (v_i, r_{ij}, v_j) \in E \) represents an edge between nodes \( v_i \) and \( v_j \), and \( r_{ij} \in R \) denotes the relation type, the forward update of an entity \( v_i \) is computed as follows~\cite{RCGN}:

\begin{equation}
h^{(l+1)}_i= \sigma (\sum_{r_{ij} \in R} \sum_{j \in N_i^r} {1 \over c_{i,r} } W_r^{(l)} h_j^{(l)} + W_0^{(l)} h_i^{(l)} )\label{eq:1}
\end{equation}

\noindent where \(h_i^{(l)}\) is the hidden state of entity \(v_{i}\) in the $l$-th layer, \(N_i^r\) is the set of neighbor indices of node $i$ under relation $R$, \(c_{i,r}\) is a normalization constant that can be either learned or fixed, and \(W_r^{(l)}\) is the weight matrix of $l$-th layer.

The full RGCN model is specified as follows: $L$ layers as defined in \eqref{eq:1} are stacked together, where the output of each layer is fed as an input to the next layer. The first layer's input is the original graph node features.

In order to perform link prediction, we use the same model introduced by \cite{RCGN}. Given a graph \(\emph{G}\), the goal is to establish the likelihood that an edge \((v_i, r_{ij}, v_j)\) belongs to the set of edges \(\emph{E}\). This is achieved by assigning a score \(f (v_i, r_{ij} ,v_j)\) to each possible edge \( (v_i, r_{ij} ,v_j)\) in the graph. The calculation of this score is achieved using an RGCN encoder with a DistMult decoder. First, each node \( v_i\) is converted by the encoder into a d-dimensional vector of real values \( e_i \in \mathbb{R}^d \). The DistMult decoder then scores each relation \((v_i, r_{ij} ,v_j)\) using a function \( f : \mathbb{R}^d  \times \emph{R}  \times \mathbb{R}^d \to \mathbb{R} \) as follows:
 
\begin{equation}
f (v_i, r_{ij} ,v_j) = e_i^T R_r e_j
\end{equation}

\noindent where \( e_i\) and \(e_j \) are $d$-dimensional vectors mapped from  \( v_i \) and \(v_j \) respectively by RGCN encoder. \( R_r \in R^{d \times d}\) is a diagonal matrix mapped from each relation $r_{ij}$ using DistMult \cite{yang2014embedding}.

\subsubsection{V2X}
The output of the RGCN generates a graph of lane centerlines for each of the input images, which is then transformed into a GeoJSON format and send to the cloud. The process includes: (1) Extraction of outputs nodes (waypoints) and edges (lane connections); (2) Coordinate Mapping where the centerlines are projected into real-world coordinates (e.g., WGS84) using localization data;
(3) GeoJSON encoding where the centerlines are stored as LineStrings with attributes like lane type and curvature.

The final step is to send the map data for each frame to the cloud via V2X (e.g., C-V2X or ITS-5G/DSRC), where multiple vehicle inputs are aggregated into a global HD map.

\subsection{Model Training}
We use nuScenes dataset~\cite{nuscenes} for training and evaluating HDMapLaneNet, as it includes Ground Truth (GT) centerline coordinates for 1000 scenes from Boston and Singapore. During training, we perform Hungarian matching between the estimated and GT lanes as described in \cite{STSU} using a matching loss
\begin{equation}
\mathcal{L}_{matching} = \mathcal{L}_{CE} + \lambda \mathcal{L}_1
\end{equation}

\noindent
where \( \mathcal{L}_{CE}\) is the detection cross-entropy loss and \( \mathcal{L}_1\) is the $\mathcal{L}_1$-norm loss on Bézier control points locations. In order to assess how good the estimated graph is compared to the GT, we consider three evaluation metrics. 

\subsubsection{Precision-Recall}
\begin{figure}[t!]
    \centering  
    \includegraphics[scale=0.33, trim= 2cm 0 0cm 0,clip]{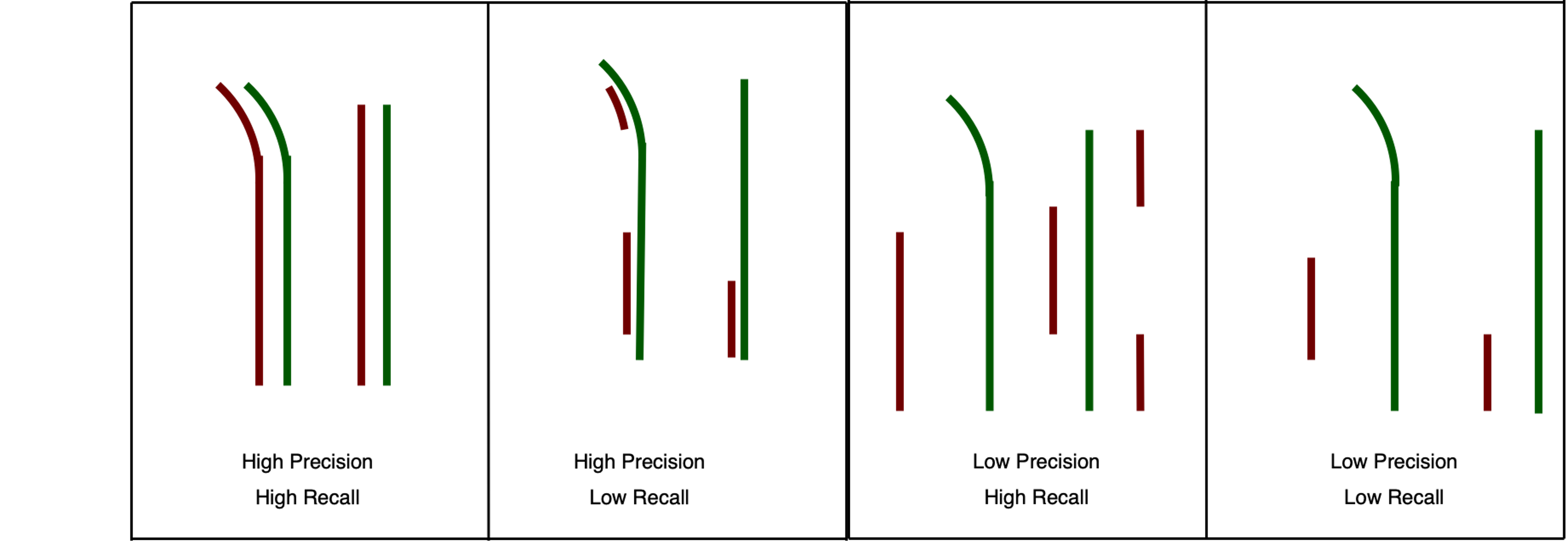}
    \caption{Illustration of the Precision-Recall Metric.}
    \label{fig:exple}
\end{figure}

The aim is how accurately generated subgraphs fit GT centerlines. The Precision-Recall metric is based on matched centerlines. After matching each estimated centerline to the GT target with the minimum \(L_1\) loss on Bézier control coefficients, we perform interpolation to have a dense representation of Bézier coefficients. If the interpolated point is close enough (compared to a threshold) to the matched GT point then it is considered as a true positive, else it is classified as a false positive. Examples of Precision-Recall are shown in Fig. \ref{fig:exple}. Red lines correspond to estimated centerlines and green ones correspond to ground truth centerlines.
 
\subsubsection{Detection Ratio}

In order to quantify how many GT centerlines are not estimated, we take into account GT centerlines that are not matched to any estimated centerline. We compute the so-called detection ratio, which is the ratio of GT centerlines that have been matched to at least one estimated centerlines over the total number of GT centerlines.

\subsubsection{Connectivity Precision-Recall}

To quantify the connectivity of the estimated centerlines compared to the GT centerlines, we measure how well the associations between the estimated centerlines are formed. For this purpose, we compute a connectivity precision-recall metric as described in STSU \cite{STSU}. Let $E$ and $I$ be, respectively, estimated incidence matrices and GT incidence, and $M(i)$ be the target index to which the $i$-th estimation is matched, and $S(n)$ be the set of estimation indices that are matched to node n. An element \(E_{ij} =1\) is a true positive if \((M(i)=M(j) | I(M(i),M(j))=1)\) and a false positive if not. In other words, if two estimated centerlines are connected according to the incidence matrix $E$, then this association can only be true if the two estimations are matched with the exact same target, or their matched GT centerlines are associated according to $I$.  In contrast, a false negative occurs when an element \(I_{m,n} = 1\) and \(	\nexists (i,j) :(( i \in S(m)) \wedge (j \in S(n))    \wedge (E_{i,j} = 1))  \). In other words, a false negative occurs when two GT centerlines are connected \(I_{m,n}=1\), and there is no estimated centerline matched with target $m$, or there is no estimated centerline matched with target $n$, or out of all pairs of estimated centerlines $(i,j)$ such that $i$ is matched with $m$ and $j$ with $n$, there is no pair that is associated according to $E$.  

\section{Implementation and Main Results}

\subsection{Implementation}

We exclusively use frontal camera images of the nuScenes dataset~\cite{nuscenes}. In order to generate GT centerline Bézier control points, we first convert centerline coordinates from real world reference frame into frontal camera reference frame using camera intrinsic and extrinsic parameters. We then re-sample these coordinates with BEV map resolution of 25cm and normalize them before extracting control points. Our model is implemented in Pytorch, with a pretrained DeepLabv3 \cite{deeplabv3} on CityScapes \cite{Cordts2016Cityscapes}. For our link predictor, we use the implemented RGCN in Pytorch Geometric. Training was performed using an HPC cluster including 4 Nvidia Tesla V100 SXM2 GPU nodes. 

\subsection{Main Results}

\begin{figure}[t]
     \centering
     \begin{subfigure}[b]{0.39999\columnwidth}
         \centering
         \includegraphics[scale=0.14, width=\columnwidth]{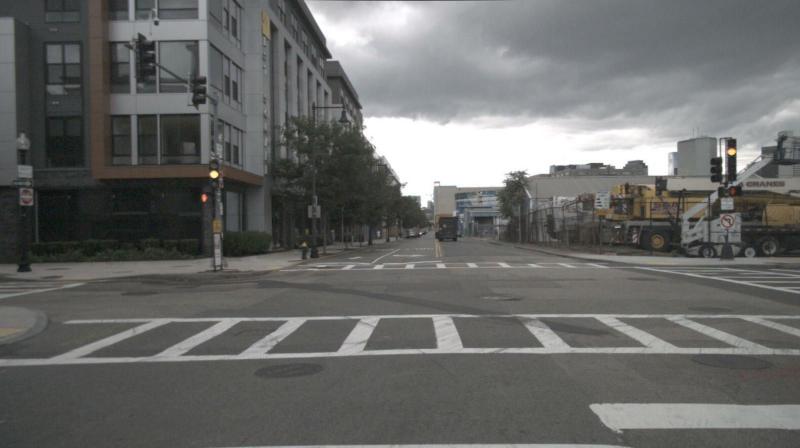}
         \label{fig:y equals x}
     \end{subfigure}
     \begin{subfigure}[b]{0.23\columnwidth}
         \centering
         \includegraphics[width=\columnwidth]{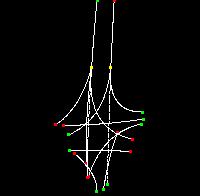}
         \label{fig:three sin x}
     \end{subfigure}
     \begin{subfigure}[b]{0.23\columnwidth}
         \centering
         \includegraphics[width=\columnwidth]{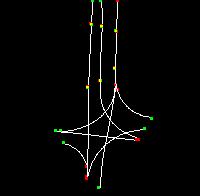}
         \label{fig:five over x}
     \end{subfigure}
     \begin{subfigure}[b]{0.39999\columnwidth}
         \centering
         \includegraphics[scale=0.14, width=\columnwidth]{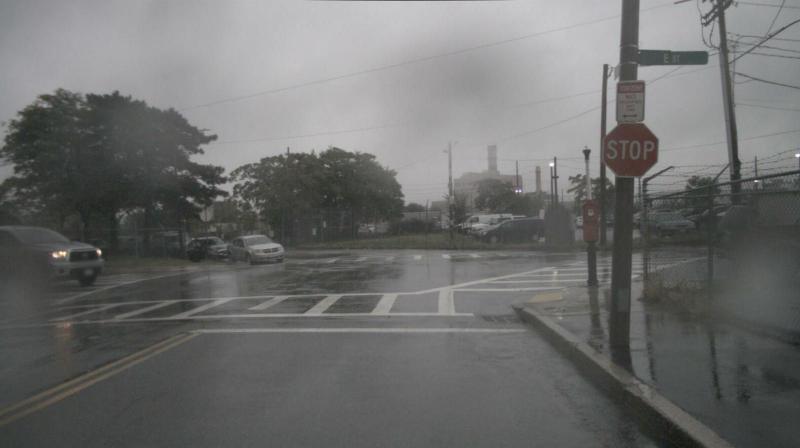}
         \label{fig:y equals x}
     \end{subfigure}
     \begin{subfigure}[b]{0.23\columnwidth}
         \centering
         \includegraphics[width=\columnwidth]{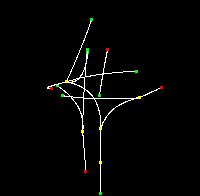}
         \label{fig:three sin x}
     \end{subfigure}
     \begin{subfigure}[b]{0.23\columnwidth}
         \centering
         \includegraphics[width=\columnwidth]{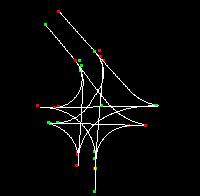}
  \label{fig:five over x}
     \end{subfigure}
    \begin{subfigure}[b]{0.39999\columnwidth}
         \centering
         \includegraphics[scale=0.14, width=\columnwidth]{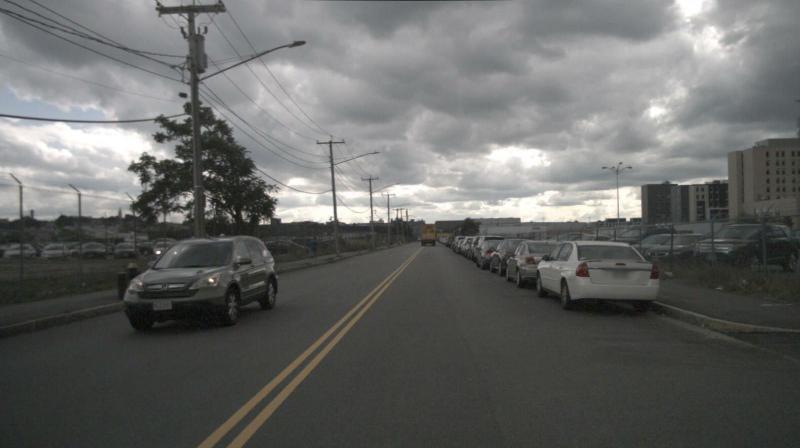}
         \label{fig:y equals x}
     \end{subfigure}
     \begin{subfigure}[b]{0.23\columnwidth}
         \centering
         \includegraphics[width=\columnwidth]{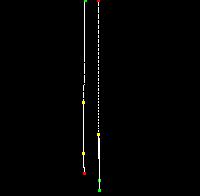}
         \label{fig:three sin x}
     \end{subfigure}
     \begin{subfigure}[b]{0.23\columnwidth}
         \centering
         \includegraphics[width=\columnwidth]{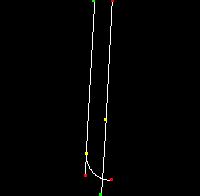}
         \label{fig:five over x}
     \end{subfigure}     
    
    \begin{subfigure}[b]{0.39999\columnwidth}
         \centering
         \includegraphics[scale=0.14, width=\columnwidth]{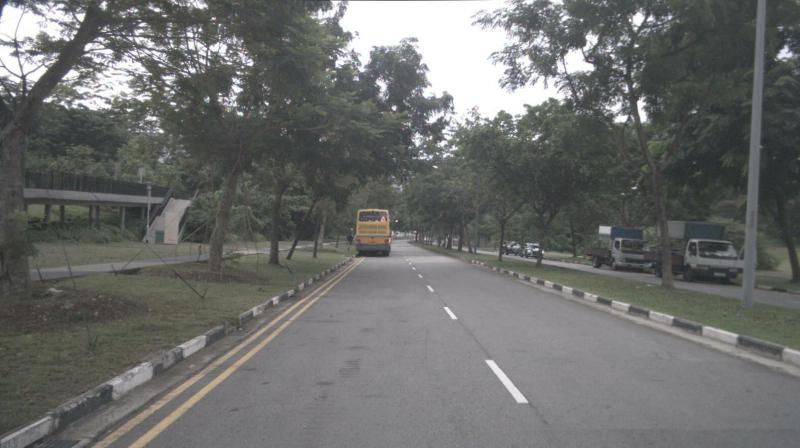}
         \caption{Raw image}
         \label{fig:y equals x}
     \end{subfigure}
     \begin{subfigure}[b]{0.23\columnwidth}
         \centering
         \includegraphics[width=\columnwidth]{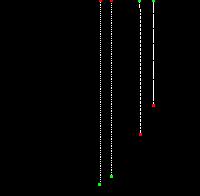}
         \caption{Ours}
         \label{fig:three sin x}
     \end{subfigure}
     \begin{subfigure}[b]{0.23\columnwidth}
         \centering
         \includegraphics[width=\columnwidth]{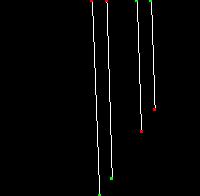}
         \caption{GT}
         \label{fig:five over x}
     \end{subfigure}     
        \caption{Qualitative results on nuScenes dataset~\cite{nuscenes}.}
        \label{fig:qualitative_nuscenes}
\end{figure}

The results achieved by our best model are shown in Table \ref{table:1}. Since our model uses the same detection head, DETR~\cite{detr}, as in STSU~\cite{STSU}, the detection ratio, detection precision, and recall metrics remain relatively similar for both models. However, HDMapLaneNet outperforms in association prediction, attributed to the proven high performance of RGCN in link prediction.  
Qualitative results of HDMapLaneNet on the nuScenes dataset are shown in Fig.~\ref{fig:qualitative_nuscenes}. Our model can predict map graphs even in challenging conditions, such as adverse weather or the presence of occlusions. It demonstrates relatively higher precision in detecting straight centerlines and accurately predicting associations between them. Results could be further improved by incorporating images from other views as well as other sensing modalities such as lidar point clouds.   

\begin{table}[t!]
\centering
\caption{HDMapLaneNet Results. D-Precision and D-Recall refer to detection metrics while C-Precision and C-Recall refer to connectivity metrics.}
\begin{tabular}{ p{1.75cm}|p{0.9cm} p{0.85cm} p{0.95cm} p{0.9cm} p{1cm}} 
 \hline
 \hline
 Method & D-Prec. & D-Rec. & D-Ratio & C-Prec. & C-Rec \\  
 \hline
 STSU & 60.7 & 54.4 & 60.6 & 60.5 & 52.2 \\ 
 HDMapLaneNet & 60.5 & 54.6 & 59.2 & \textbf{75.9} & \textbf{67.1}\\ 
 \hline
 \hline
\end{tabular}
\label{table:1}
\end{table}

\section{Conclusion \& Future work}
\hspace{\parindent}
This paper presented \textit{HDMapLaneNet}, a novel framework for collaboratively generating a localized geometric layer of HD maps using a V2X communication interface. The proposed method enables individual vehicles to extract lane centerlines from front-facing camera images, structure them into a graph representation, and transmit the data for global HD map aggregation. Preliminary results on the nuScenes dataset indicate that the framework outperforms a state-of-the-art method in association prediction, validating its effectiveness for accurate lane mapping. Future work will focus on using several cameras for larger field of view and on efficiently merging the graph representations from individual vehicles to generate and validate the global geometric layer of the HD map.

\section*{Acknowledgments}

This work was supported in part by the Fonds National de la Recherche of Luxembourg (FNR) through the Project FNR AUTOMAP under Grant BRIDGES2020/IS/15354216, and in part by the Industrial Partnership between the Interdisciplinary Center for Security Reliability and Trust (SnT) of the University of Luxembourg and Luminar Technologies (Civil Maps).

\bibliographystyle{IEEEtran}
\bibliography{IEEEabrv,bibliography/references}

\end{document}